\documentclass{article}

\usepackage{arxiv}

\usepackage[utf8]{inputenc} % allow utf-8 input
\usepackage[T1]{fontenc}    % use 8-bit T1 fonts
\usepackage{hyperref}       % hyperlinks
\usepackage{url}            % simple URL typesetting
\usepackage{booktabs}       % professional-quality tables
\usepackage{amsfonts}       % blackboard math symbols
\usepackage{nicefrac}       % compact symbols for 1/2, etc.
\usepackage{microtype}      % microtypography
\usepackage{lipsum}
\usepackage{fancyhdr}       % header
\usepackage{graphicx}       % graphics
\usepackage{caption}
\usepackage{subcaption}
\usepackage{float}

\title{VisualEnv: visual Gym environments with Blender}

\author{
  Andrea Scorsoglio \\
  Space Systems Engineering Laboratory \\
  University of Arizona \\
  Tucson, AZ\\
  \texttt{andreascorsoglio@email.arizona.edu} \\
   \And
  Roberto Furfaro \\
  Space Systems Engineering Laboratory \\
  University of Arizona \\
  Tucson, AZ\\
  \texttt{robertof@email.arizona.edu} \\
}

\begin{document}
\maketitle

\begin{abstract}
In this paper VisualEnv, a new tool for creating visual environment for reinforcement learning is introduced. It is the product of an integration of an open-source modelling and rendering software, Blender, and a python module used to generate environment model for simulation, OpenAI Gym. VisualEnv allows the user to create custom environments with photorealistic rendering capabilities and full integration with python. The framework is described and tested on a series of example problems that showcase its features for training reinforcement learning agents.
\end{abstract}

\section{Introduction}
Interest in Reinforcement Learning (RL) has been on the rise over the past decade. Among the most advanced algorithms available are Deep Q-Learning (DQN\cite{mnih2013playing}), Deep Deterministic Policy Gradient (DDPG\cite{lillicrap2015continuous}), Asynchronous Advantage Actor-Critic (A3C\cite{mnih2016asynchronous}), Thrust Region Policy Optimization (TRPO\cite{schulman2015trust}) and Proximal Policy Optimization (PPO\cite{schulman2017proximal}). They have proven to excel in many robotic motion tasks \cite{Kober:2013:RL_robotics_survey,Nakanishi:2004:adaptation_biped_loco,Smart:2002:RL_mobile_robots,Peters:2006:PG_methods_robotics} as well as games \cite{silver2016mastering,kempka2016vizdoom,vinyals2017starcraft}. More recently, reinforcement learning has been used in a variety of applications, from autonomous vehicle navigation \cite{kiran2021deep,shalev2016safe} to spacecraft guidance tasks \cite{scorsoglio2019actor,scorsoglio2020image,furfaro2020adaptive,scorsoglio2020safe}. Among these use scenarios, the most challenging tasks for RL are typically the ones where the agent has access to optical sensor to navigate the surrounding world. Extracting information from images is the ultimate goal of reinforcement learning as optical sensor are relatively inexpensive and readily available. Moreover, vision is the primary tool we humans have to navigate our surroundings so it is the natural direction to follow for humanoid level of interaction with the world for machines.  Although there are some examples of simulated visual environments, they are all designed for very specific tasks that cannot be easily adapted for other applications (ATARI \cite{bellemare2013arcade}, Doom \cite{kempka2016vizdoom}, Starcraft \cite{vinyals2017starcraft}, Dota \cite{berner2019dota}). Many of them rely on a specific game engine that cannot extend to other applications. For this reason, these tools are used for benchmarking purposes more than real world applications. An example of custom environment creator for RL application is Unity Machine Learning Agents, which allows the user to train RL agent using the built in game engine. It is very powerful as it has access to a variety of tools for vector observations, such as raycasting, as well as real time visualization and parallelization. Although powerful, it was created with the intent of easily implement AI agents in video games, not for scientific research. This hinders its application to more realistic usage scenarios.

With this paper, a new tool for creating custom visual environment for reinforcement learning is introduced. VisualEnv is a sandbox simulator that offers complete control over the environment generation and interactions. The tool is built around two open-source pieces of software: Blender \cite{blender} and OpenAI Gym \cite{brockman2016openai}. Blender is a modelling, animation and rendering software that is usually used for production quality computer graphics workflows. It offers a wide range of tools for modelling, sculpting, shading, texturing, lighting and rendering. Moreover, it interfaces easily with python through an API. OpenAI Gym on the other hand is a powerful environment builder written in python and widely used to train RL agents. VisualEnv harnesses the power of both to create a standalone package that can be used to generate custom visual environment in python. These can then be used to train RL algorithm or perform other types of real time simulation tasks. The user has control over every aspect of the rendering and animation of the environment. The environment and the actor can be created using a pletora of photorealistic shaders and materials. The environment dynamics can be controlled by the gym environment that then interfaces with blender through the API, moving and modifying the actors in the scene. The rendered observations can then be used for a variety of applications. The power of the frameworks resides in the ease of implementation and capabilities suitable for online-applications.

In this paper, the framework is presented and described in details. Its capabilities are then showcased on three test environments. Specifically, it is first tested on a visual version of the common OpenAI gym CartPole-v0 environment where the only observation given to the agent is a front view of the cart-pole system. It is then tested on an environment where the agent has to hover over a randomly positioned target using a camera pointed towards the ground. Finally, it is tested on the Goalie environment, where the agent has to intercept a ball thrown towards it before it crosses the goal line, with the only observation available being the camera view of the scene from the agent perspective.

\section{Background}
In RL tasks, an \textit{agent} is considered to be situated in a simulated environment. The environment is generally described as a partially observable Markov decision process (POMDP) which has a \textit{state} and \textit{action} spaces that encompass all the possible states and actions. The agent selects an action at each step and observes the evolution of the environment. From the environment it receives an \textit{observation} and a \textit{reward}. The goal of the agent is to maximize the reward obtained as it interacts with the environment. With the introduction of OpenAI Gym \cite{brockman2016openai}, the creation of environments for reinforcement learning has become more streamlined that in the past. It is mostly used in two ways: either to benchmark RL codes on consistent pre-made environments, or to create custom environments with a common structure that can then be used to train RL agents. The tool allows the user to create environment with continuous or discrete state and actions. The observation can also be an image seen as a hypercube with the dimensions of the image. The structure is described in details in the OpenAI gym documentation\footnote{OpenAI Gym documentation: \url{https://gym.openai.com/docs/}.} and is omitted here. The user should be familiar with it in order to be able to use VisualEnv. The VisualEnv environment is created as a Gym environment with added rendering capabilities provided by Blender.

Blender \cite{blender} is an open-source modelling, animation and rendering software. Its tools allow the user to model complex shapes using a variety of techniques. One can create landscapes through terrain displacement using digital elevation models, as well as populate the environment with any 3D objects. Complex geometries can be generated within the program using the numerous modelling tools available or imported from other software (imported as  .obj, .fbx, .3ds, .stl). Each object can be given a material, which is controlled through a shader editor that allows great freedom on its physical properties and its interaction with the light. It is also possible to control the lighting of the scene using a variety of elements such as spotlights, sun-like sources, as well as emissive objects that allow for extremely realistic lighting conditions. All the elements introduced above allow the user to create photo-realistic renderings of the scene. The render options are controlled by the camera properties. The camera focal length, the resolution and the sensor size can all be adjusted. The renderer takes into account the physics of both the materials and the camera to computes multiple light bounces coming from all the sources in the scene as they interact with the materials to obtain physically accurate results.

The reason why this tool is so appealing for reinforcement learning application is its affinity with python. It runs natively in python with most of the libraries written in C++. This architecture makes Blender suitable for python integration without sacrificing performance.

\section{VisualEnv motivation and goals}
VisualEnv is a tool that enables the creation of custom visual environments through the integration of Blender and Gym. The motivation for VisualEnv comes from the need of a scalable and general framework for visual environment for RL. There is a multitude of examples of visual environment used in RL. The Arcade Learning Environment (ALE)\cite{bellemare2013arcade} is a library for ATARI 2600 retro games that has been used extensively as a benchmark for vision based RL agents. More recently, a series of game engines have been used to create RL agents capable of beating professional human players (Doom \cite{kempka2016vizdoom}, Starcraft \cite{vinyals2017starcraft}, Dota \cite{berner2019dota}). What all these environments have in common is the fact that they are self-contained applications that can only be used as benchmarking tool. There are currently no open-source solutions for developing custom visual RL environments, with the exception of Unity Machine Learning Agents. The package is very powerful, but it lacks control over the rendering capabilities and is more oriented towards implementing AI agents into game environments than providing a research platform for phisically accurate applications. VisualEnv on the contrary, uses the powerful cycles renderer within blender to produce physically accurate renders, and the native python API makes it possible to interface the environments with any available or custom RL package. The goal is to inherit the powerful attributes that make Gym ubiquitous and the de-facto standard when it comes to simulating RL environments. This translates into the following goals:
\begin{itemize}
    \item Make the package easy to use, with minimal intervention from the user.
    \item Make the package general enough so that it can adapt to most of the usage cases.
    \item Maintain compatibility with most of the available libraries for training RL agents.
\end{itemize}
This is achieved by keeping most of the standard Gym API, introducing helper elements that allow for a seamless incorporation of visual observations into the environments.

\section{VisualEnv custom environments}

\begin{figure} [H]
    \centering
    \includegraphics[width=0.8\textwidth]{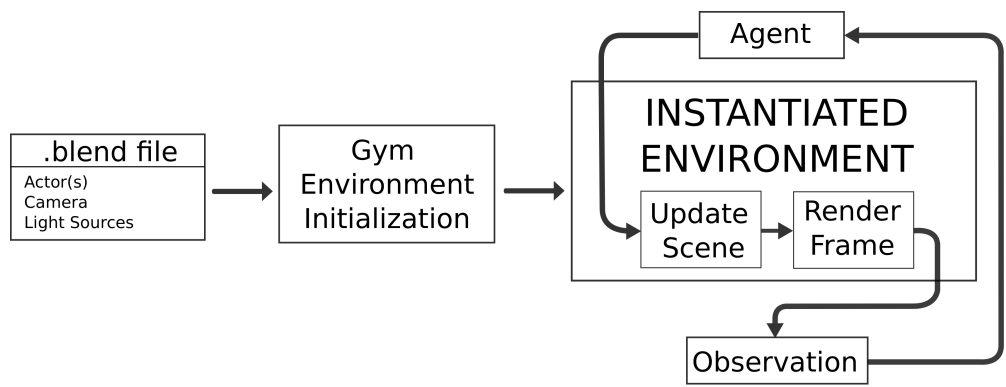}
    \caption{VisualEnv Architecture}
    \label{fig:vb}
\end{figure}

The VisualEnv essentially inherits the structure of an OpenAI Gym environment with some key additions. Figure \ref{fig:vb} describes the high level architecture of the system when used to train RL agents.

The user has to produce two major components: the blender file where the environment and the actors reside, and the gym environment functions, responsible of controlling the dynamics of the environment. The two components are interfaced by importing the blender file within the gym environment. 
To populate the scene the user can rely on the tools offered by Blender. Objects can be modelled in blender through the GUI, created using the python API or imported from other .blend files or from other software. The possibilities are endless and there is no limit on the scene complexity. The user also has a lot of freedom on the design of the materials as they interact with the light sources. Some examples created using blender are given the reference, where it was used to simulate a physically accurate lunar surface \cite{scorsoglio2020image,scorsoglio2020safe}. The user is not restricted to use a single camera, multiple cameras can be used simultaneously if the need should arise.  All the objects in the scene are then automatically available within the main gym environment script. The scene (actors, light sources, cameras...) can be manipulated using the native blender python API. This allows for completely dynamic environments where every attribute can be changed at each time step. Among other options, the user can change the position and rotation of the objects as well as change their appearances (both shape and material). The scene can then be rendered out using any of the cameras present in the scene. The parameters of the cameras can be assigned beforehand in the blender file or through the API within the environment script. The rendered image(s) (RGB or grayscale) are then available to the environment as pixel arrays. They can then be manipulated further or passed as-is to the agent.

The core functionality of VisualEnv is to be able to interact with a completely customizable visual environment that is updated in real time. This is pivotal for online application such as reinforcement learning. The framework here presented allows for fast interactions between the renderer and the training algorithm while maintaing a high level of flexibility. The renderer can work both on CPU or GPU (via enabling the appropriate toggle either through the GUI or the python API). The possibility of using the GPU can greatly increase performance in case the rendered images become large.

\section{Test environments}
Although the focus of this paper is to present the environment creation toolbox, some test environments were created and used to train an agent using reinforcement learning. The agents are trained using the RL baselines implemented in PyTorch (Stable Baselines3 \cite{stable-baselines3}). The test problems are reported below.

\subsection{Visual CartPole}
\begin{figure} [H]
     \centering
     \begin{subfigure}[b]{0.4\textwidth}
         \centering
         \includegraphics[width=\textwidth]{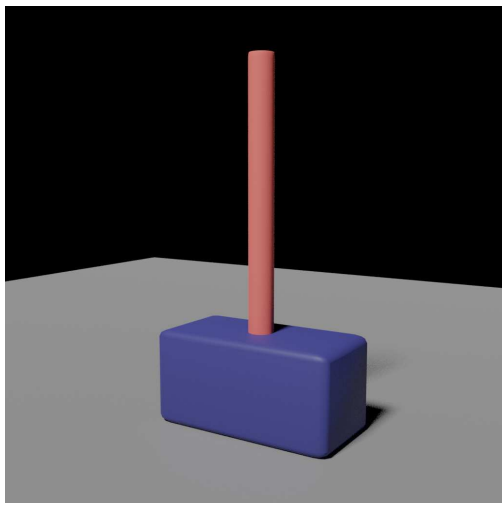}
         \caption{High resolution view of the environment}
         \label{fig:cart_ext}
     \end{subfigure}
     \hfill
     \begin{subfigure}[b]{0.4\textwidth}
         \centering
         \includegraphics[width=\textwidth]{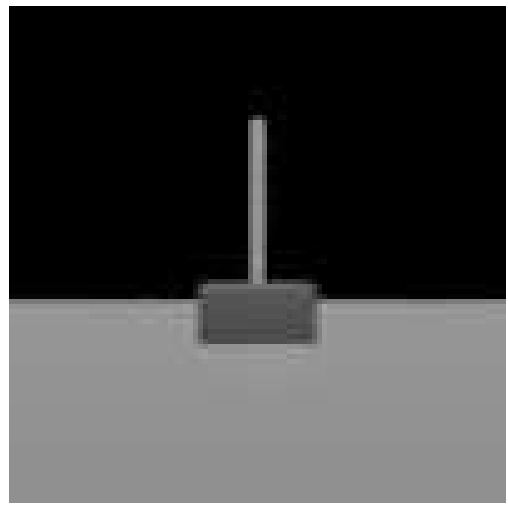}
         \caption{Grayscale observation fed to the network}
         \label{fig:cart_obs}
     \end{subfigure}
        \caption{Visual Cartpole}
        \label{fig:Vcart}
\end{figure}

A visual version of the CartPole environment from OpenAI Gym (CartPole-v0). The environment is composed by a cart that is free to move horizontally in one dimension. A pole is attached on top of the cart and is free to rotate around a hinge. The goal is to balance the inverted pole on top of the cart. The only possible action is a finite impulse applied to the cart in either direction. The episodes is terminated whenever the pole falls below a certain threshold, the cart goes out of bounds or the maximum number of time-steps is reached. The game is won if the agent manages to balance the pole for more than 100 steps. The reward is 1 for each time-step until the game is reset or won. The observation in this case is the grayscale image of the cart-pole system as seen by a fixed camera facing the cart. A high resolution representation of the environment and a grayscale observation image are shown in Figure \ref{fig:Vcart}.

\subsection{Hover2D}
\begin{figure} [H]
     \centering
     \begin{subfigure}[b]{0.4\textwidth}
         \centering
         \includegraphics[width=\textwidth]{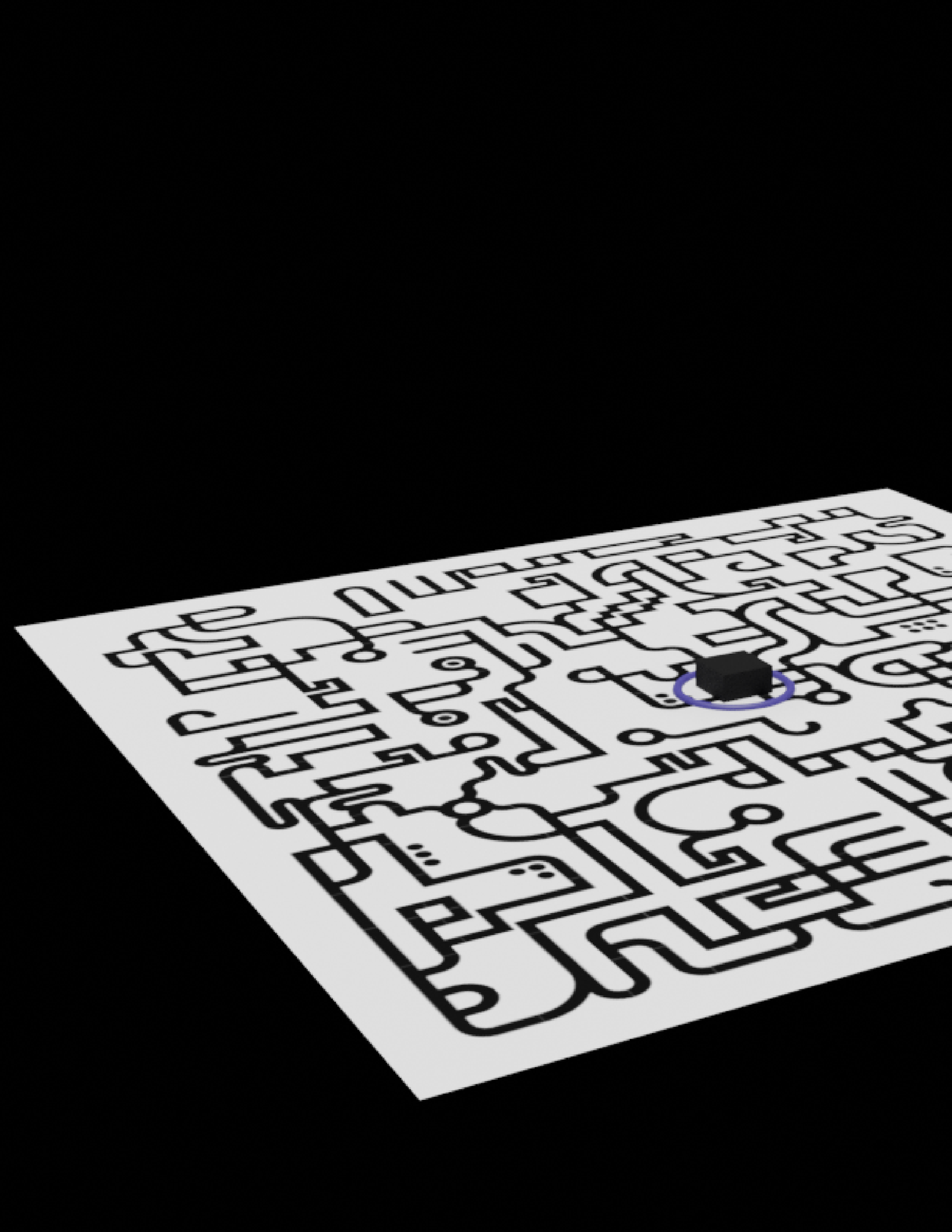}
         \caption{High resolution view of the environment}
         \label{fig:hover_ext}
     \end{subfigure}
     \hfill
     \begin{subfigure}[b]{0.4\textwidth}
         \centering
         \includegraphics[width=\textwidth]{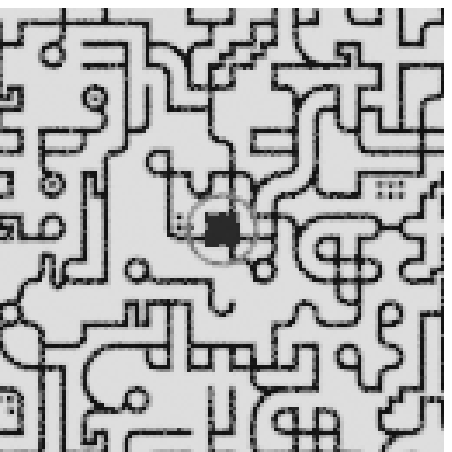}
         \caption{Grayscale observation fed to the network}
         \label{fig:hover_obs}
     \end{subfigure}
        \caption{Hover2D}
        \label{fig:Hover}
\end{figure}

In this environment the agent must hover over a target using pulses in two dimensions (up, down, left, right). The agent is constrained to move on an horizontal plane, parallel to the ground. The camera is pointed downward, perpendicular to the ground. The ground is textured with a non-repeating pattern to avoid indetermination. The target is a black cube randomly positioned on the ground at the beginning of each episode. The agent position is also initialized at random at the beginning of each episode. The goal is to hover within two meters of the target as indicated by a the circular shape around the target. Moreover, a sphere follows the horizontal movement of the agent on the ground plane to keep track of the camera bore-sight direction. The episodes continue until the agent achieves the goal, moves out of bounds, or the maximum number of time-steps is reached. The agent receives a negative -20 reward if if goes out of bounds, a reward of 10 if it reaches the maximum number of timesteps without going out of bounds and a reward of 20 if it reaches the target. A representation of the environment is given in Figure \ref{fig:Hover}.

\subsection{Goalie}
\begin{figure} [H]
     \centering
     \begin{subfigure}[b]{0.4\textwidth}
         \centering
         \includegraphics[width=\textwidth]{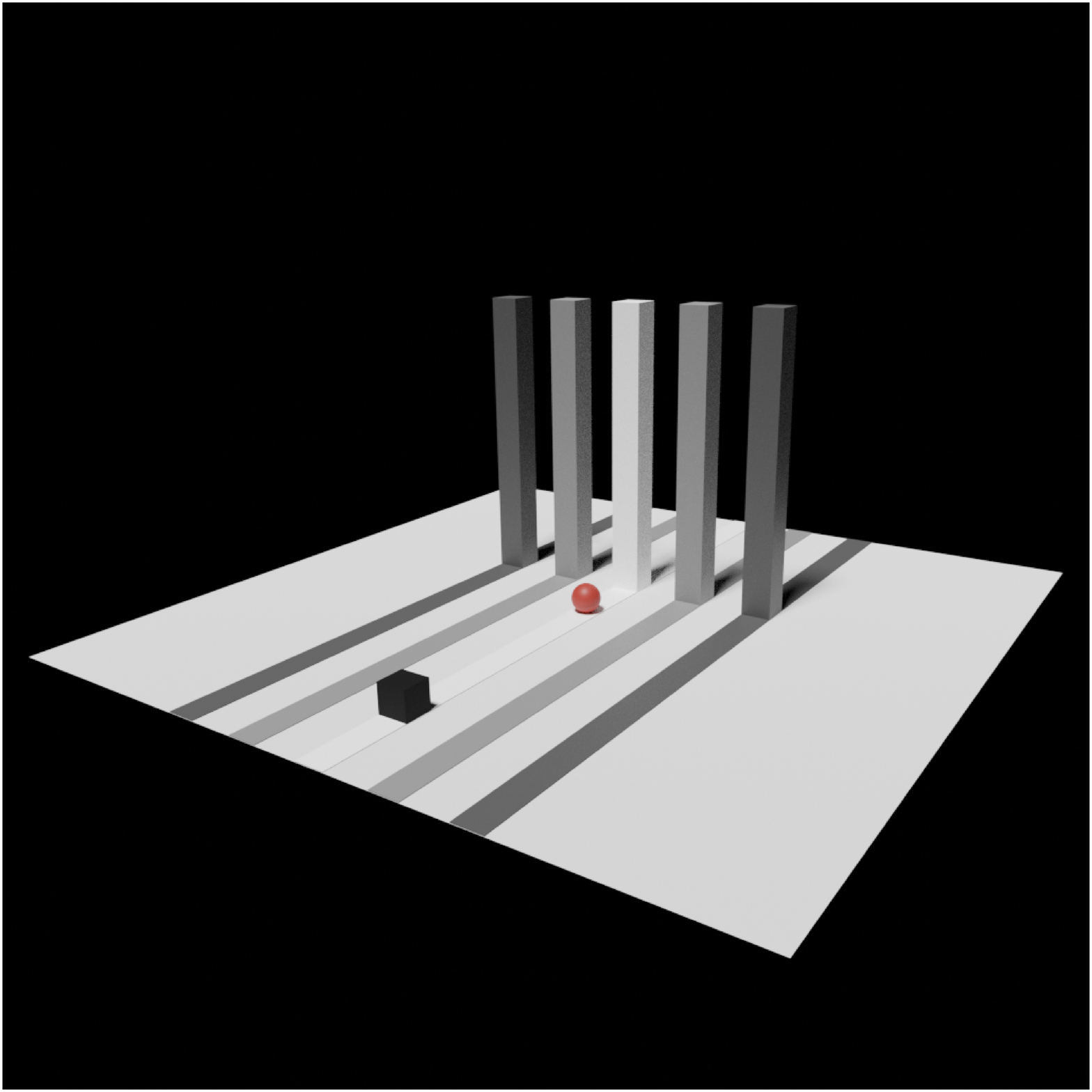}
         \caption{High resolution view of the environment}
         \label{fig:goalie_ext}
     \end{subfigure}
     \hfill
     \begin{subfigure}[b]{0.4\textwidth}
         \centering
         \includegraphics[width=\textwidth]{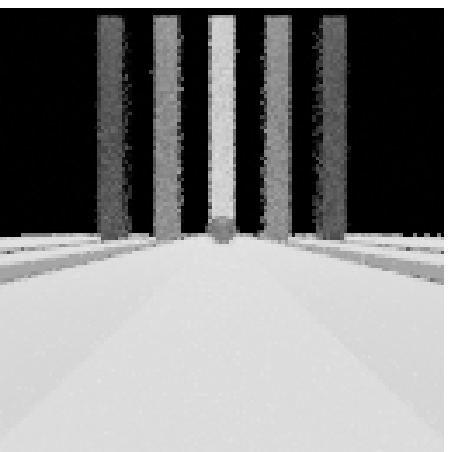}
         \caption{Grayscale observation fed to the network}
         \label{fig:goalie_obs}
     \end{subfigure}
        \caption{Goalie}
        \label{fig:goalie}
\end{figure}

In this environment, the agent is a virtual goalie whose task is to catch a ball moving towards the goal line. The ball moves at a constant speed along a direction randomly sampled between a minimum and maximum angle with respect to the field axis. The goalie can only move left or right of exactly one unit per time-step. The task of the goalie is to be within 1 unit from the ball when it crosses the goal line. The agent receives a negative reward of -10 if it goes out of bounds or if the ball reaches the goal line, otherwise if the goalie catches the ball, it receives a reward of 10. An episode ends either when the goalie catches the ball or when the ball crosses the goal line or the goalie goes out of bounds. The field is colored with a non-repeating pattern to avoid ambiguities. An array of columns with different colors is also present at the end of the field for the same reason. A representation of the environment is given in Figure \ref{fig:goalie}.

\subsection{Results}
\begin{figure} [H]
    \centering
    \includegraphics[width=0.6\textwidth]{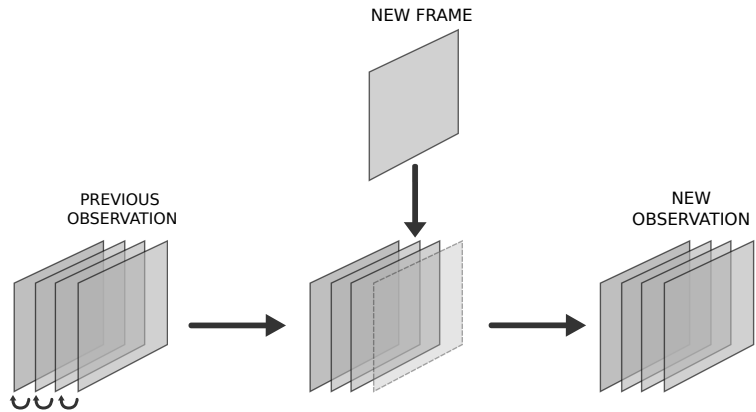}
    \caption{Stacking process}
    \label{fig:stack}
\end{figure}

\begin{figure} [H]
     \centering
     \begin{subfigure}[b]{0.4\textwidth}
         \centering
         \includegraphics[width=\textwidth]{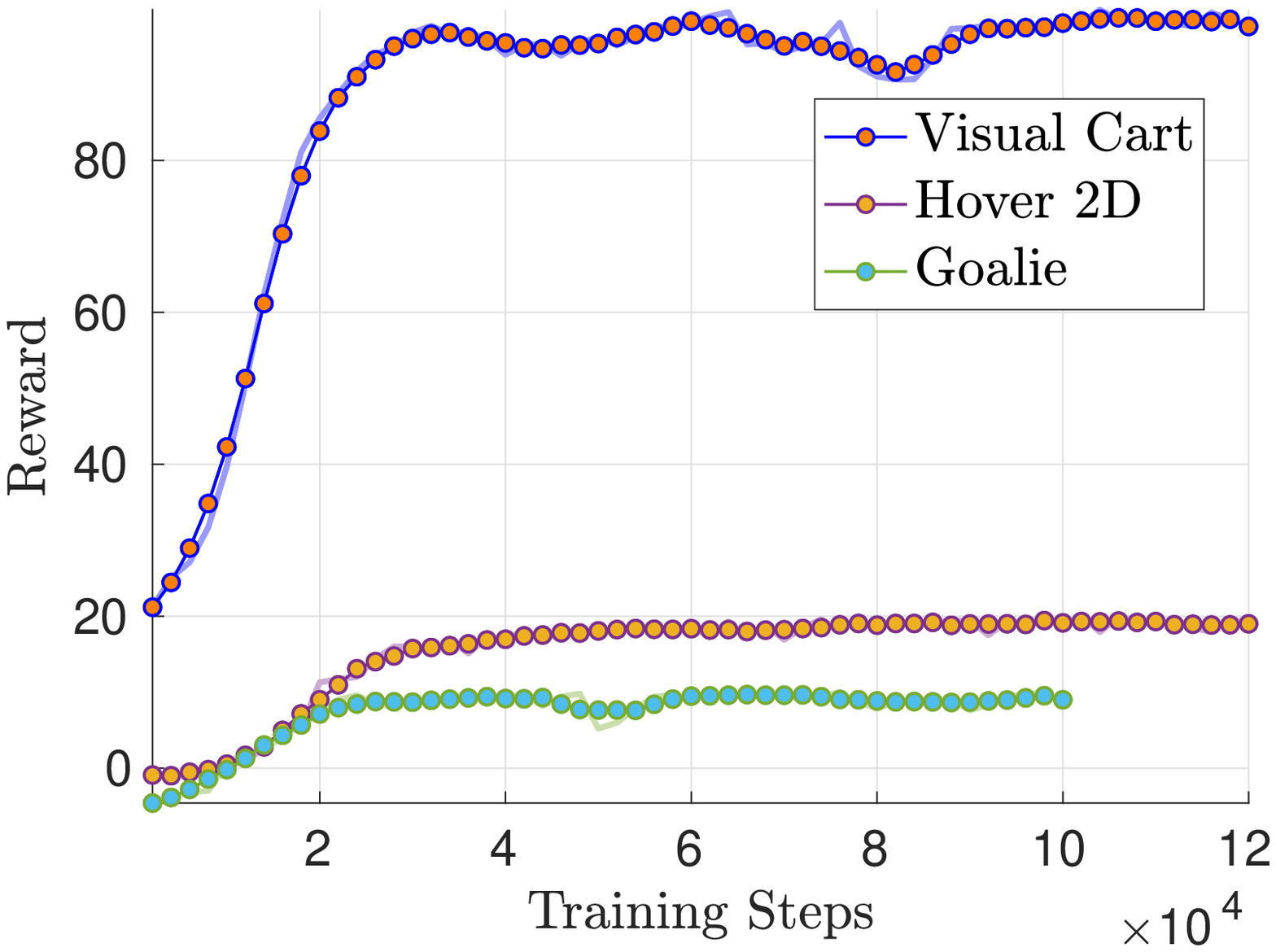}
         \caption{Mean reward on roll-outs}
         \label{fig:rew_roll}
     \end{subfigure}
     \hfill
     \begin{subfigure}[b]{0.4\textwidth}
         \centering
         \includegraphics[width=\textwidth]{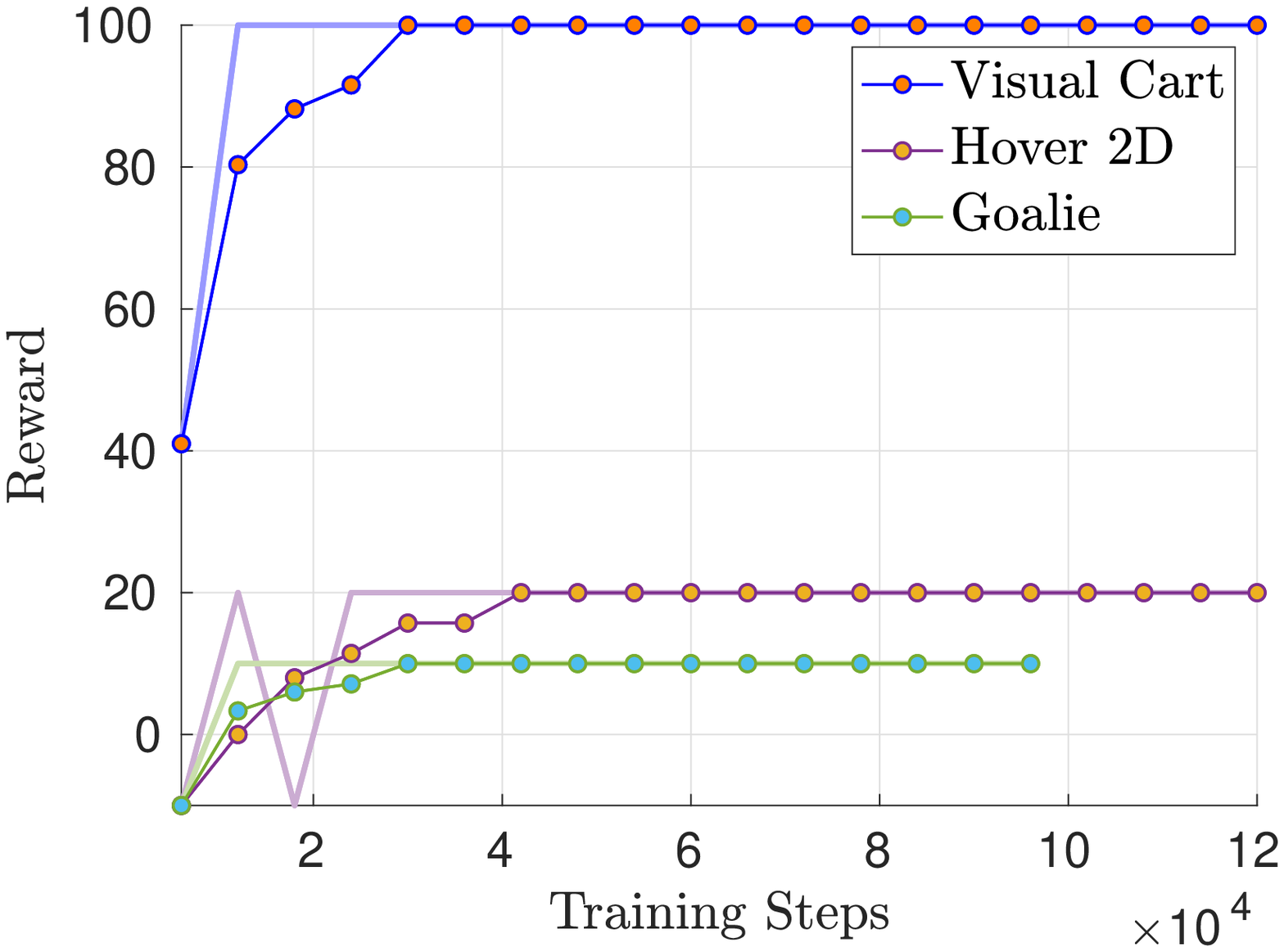}
         \caption{Mean reward on evaluation environment}
         \label{fig:rew_eval}
     \end{subfigure}
        \caption{Training profiles}
        \label{fig:rew}
\end{figure}

\begin{figure} [H]
     \centering
     \begin{subfigure}[b]{0.4\textwidth}
         \centering
         \includegraphics[width=\textwidth]{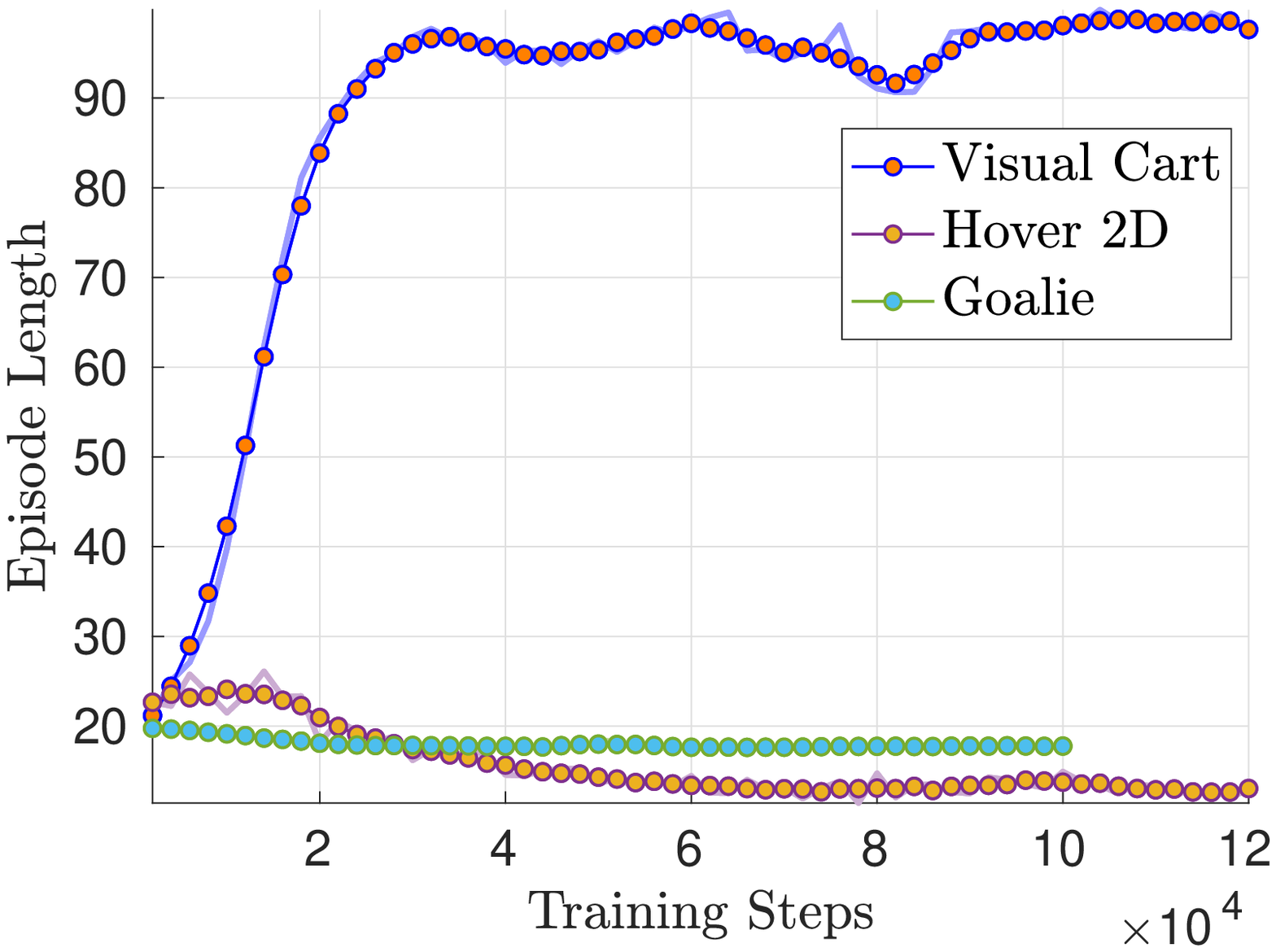}
         \caption{Mean episode length on roll-outs}
         \label{fig:len_roll}
     \end{subfigure}
     \hfill
     \begin{subfigure}[b]{0.45\textwidth}
         \centering
         \includegraphics[width=\textwidth]{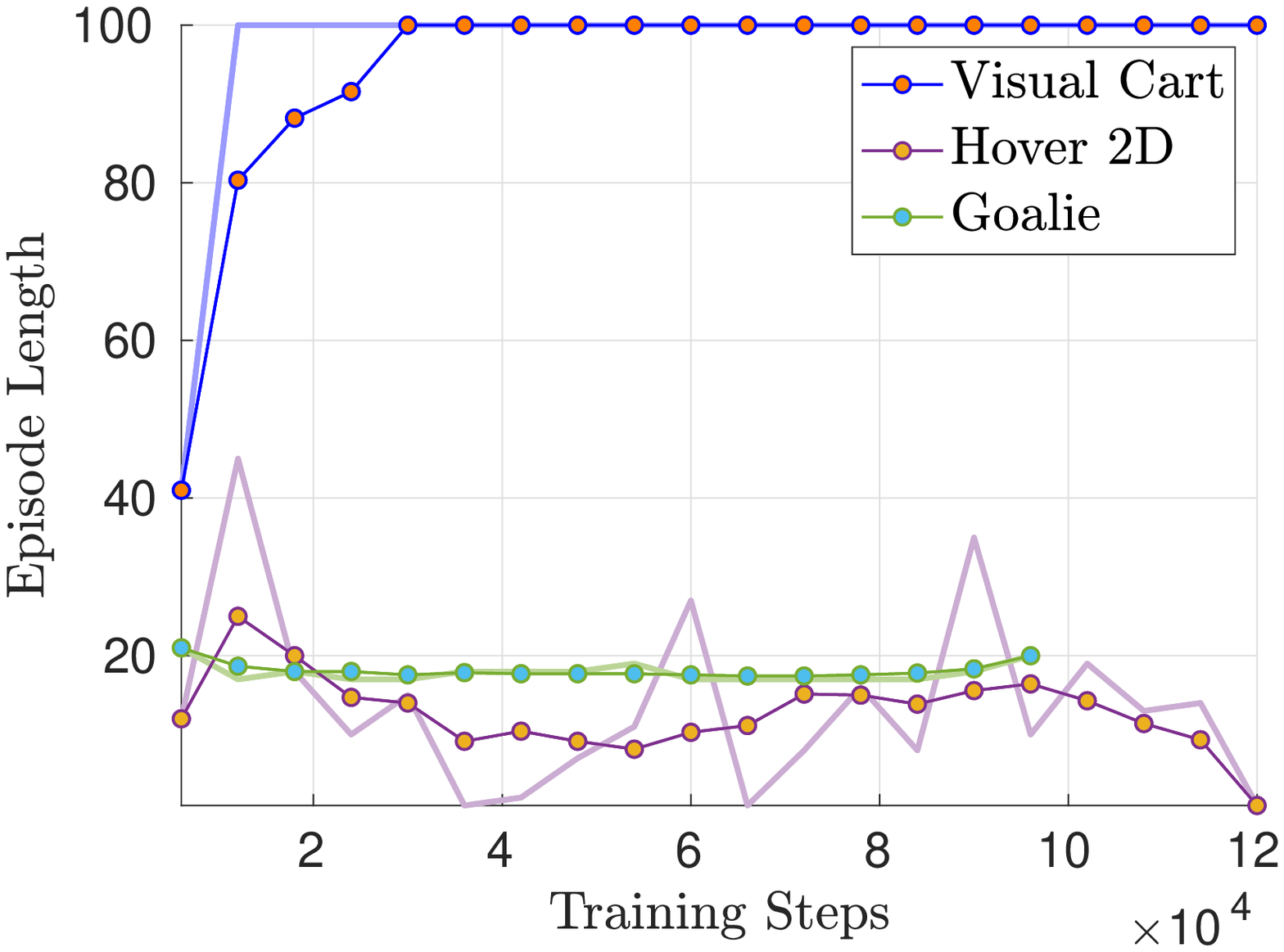}
         \caption{Episode length on evaluation environment}
         \label{fig:len_eval}
     \end{subfigure}
        \caption{Episode length}
        \label{fig:len}
\end{figure}

\begin{figure} [H]
    \centering
    \includegraphics[width=0.45\textwidth]{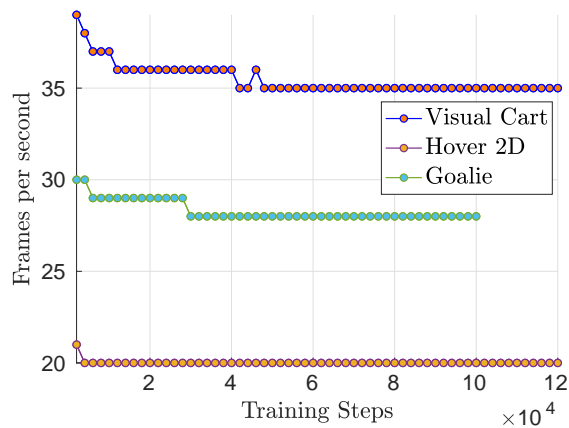}
    \caption{Samples per second during training}
    \label{fig:frames}
\end{figure}

An agent was trained in each environment using Proximal Policy Optimization (PPO\cite{schulman2017proximal}), which is known for its stability and performance. The network is the default Convolutional Neural Network (CNN) policy from Stable Baselines3. The grayscale images generated by the renderer are stacked in a 3D tensor of depth 4 before they are fed to the network. This is inspired by the ATARI benchmark problems \cite{bellemare2013arcade}, which allows the network to have partial information about the evolution of the environment without needing recurrent neural networks. Figure \ref{fig:stack} describes the staking process. All these functionalities are contained in the library and are readily available to the user.
The agents were trained for a fixed number of steps. An evaluation environment was created alongside the rollout environment and used to track the progress of learning. The model with the best validation reward was then used to test the method. Figure \ref{fig:rew} shows the mean reward during training on roll-outs and validation environments. Figure \ref{fig:len} shows the average episode length again, on roll-outs and validation environments. Finally, Figure \ref{fig:frames} shows the number of samples per second that trainer receives. It should be noted that a step comprehends not only the render time but also the overhead needed for simulating the environment dynamics. The average render time in all the cases was always observed to be below 0.03 seconds. The resolution of the observation is $100 \times 100$ in all cases and was chosen because it represents a good balance between detail and render speed.
% rew

\section{Conclusion}
VisualEnv is a tool that was born to answer to the need for a scalable and flexible environment creator for reinforcement learning with rendering capabilities. The tool is developed on top of OpenAI Gym and Blender and delivers an easy to use API for the creation of customized dynamic environments with visual observation capabilities. The user can develop an environment within the Blender GUI using all the available tools for modelling, shading and illumination. The environment can then be used to render images of the scene at run time. The ease of implementation together with physically accurate renderer allows for online realistic modelling of complex dynamic environment that can be used for simulation and training of reinforcement learning agents.

\section*{Acknowledgments and Funding}
This project was carried out and funded by the Space Systems Engineering Laboratory (\href{https://www.ssel.arizona.edu/}{https://www.ssel.arizona.edu/}) at University of Arizona.

%Bibliography
\bibliographystyle{unsrt}  
\bibliography{ms}

\end{document}